\documentclass{article}

\PassOptionsToPackage{numbers, compress}{natbib}

\usepackage[final]{neurips_2020}

\usepackage[utf8]{inputenc}
\usepackage[T1]{fontenc}
\usepackage{hyperref}
\usepackage{url}
\usepackage{booktabs}
\usepackage{amsfonts}
\usepackage{amsmath}
\usepackage{nicefrac}
\usepackage{microtype}
\usepackage{graphicx}
\usepackage{xcolor}
\usepackage{pifont}
\usepackage{caption}
\usepackage{subcaption}
\usepackage{enumitem}
\usepackage{textcomp}
\usepackage{nicefrac}
\usepackage[version=4]{mhchem}
\usepackage[nobottomtitles]{titlesec}
\usepackage{multirow}
\usepackage{siunitx}
\DeclareMathOperator{\Tr}{Tr}

\title{Instance Selection for GANs}

\author{
  Terrance DeVries \\
  University of Guelph\\
  Vector Institute\\
   \And
   Michal Drozdzal \\
   Facebook AI Research \\
   \And
   Graham W. Taylor \\
   University of Guelph \\
   Vector Institute
}
\begin{document}

\maketitle

\begin{abstract}
Recent advances in Generative Adversarial Networks (GANs) have led to their widespread adoption for the purposes of generating high quality synthetic imagery. While capable of generating photo-realistic images, these models often produce unrealistic samples which fall outside of the data manifold. Several recently proposed techniques attempt to avoid spurious samples, either by rejecting them after generation, or by truncating the model's latent space. While effective, these methods are inefficient, as a large fraction of training time and model capacity are dedicated towards samples that will ultimately go unused. In this work we propose a novel approach to improve sample quality: altering the training dataset via instance selection before model training has taken place. By refining the empirical data distribution before training, we redirect model capacity towards high-density regions, which ultimately improves sample fidelity, lowers model capacity requirements, and significantly reduces training time. Code is available at \url{https://github.com/uoguelph-mlrg/instance_selection_for_gans}.
\end{abstract}

\section{Introduction}
Recent advances in Generative Adversarial Networks (GANs) have enabled these models to be considered a tool of choice for vision synthesis tasks that demand high fidelity outputs, such as image and video generation~\citep{clark2019efficient,karras2019analyzing}, image editing~\citep{zhu2020domain}, inpainting~\citep{yu2018generative}, and superresolution~\citep{wang2018esrgan}. However, when sampling from a trained GAN model, outputs may be unrealistic just as often as they appear photo-realistic.

GANs fit a model to a data distribution with the help of a discriminator network.
Low quality samples produced by these models are often attributed to poor modeling of the low-density regions of the data manifold  \citep{karras2019style}. The majority of current techniques attempt to eliminate low quality samples after the model is trained, either by changing the model distribution by truncating the latent space \citep{brock2018large,karras2019style} or by performing some form of rejection sampling using a trained discriminator to inform the rejection process \citep{Azadi2019DiscriminatorRS,Che2020YourGI,Turner2018MetropolisHastingsGA}. Nevertheless, these methods are inefficient with respect to model capacity and training time, since much of the capacity and optimization efforts dedicated to representing the sparse regions of the data manifold are wasted. 

In this paper, we analyze the use of instance selection \citep{olvera2010review} in the generative setting. We address the problem of uneven model sample quality before GAN model training has begun, rather than after it has finished. We note that dataset collection is a noisy process, and that many of the currently used datasets for generative model training and evaluation were not purposely created for this task. Thus, through a dataset curation step, we remove low density regions from the data manifold prior to model optimization and show that this \emph{direct} dataset intervention (1) improves overall image sample quality in exchange for some reduction in diversity, (2) lowers model capacity requirements, and (3) reduces training time.
To remove the sparsest parts of the image manifold, images are first projected into an embedding space of perceptually meaningful representations. A scoring function is then fit to asses the manifold density in the neighbourhood of each embedded data point in the dataset. Finally, data points with the lowest manifold density scores are removed from the dataset. In our experiments, we evaluate a variety of image embeddings and scoring functions, observing that Inceptionv3 and Gaussian likelihood are well suited for the respective roles. Overall, we make the following contributions:
\begin{itemize}
    \setlength{\itemsep}{2pt}
    \setlength{\parskip}{0pt}
    \setlength{\parsep}{0pt}
    \item We propose dataset curation via instance selection to improve the output quality of GANs.
    \item We show that the manifold density in the perceptual embedding space of a given dataset is predictive of GAN performance, and therefore a good scoring function for instance selection.
    \item We demonstrate the \emph{model capacity savings} of instance selection by achieving state-of-the-art performance (in terms of FID) on 64 \texttimes\ 64 resolution ImageNet generation using a Self-Attention GAN with \nicefrac{1}{2} the amount of trainable parameters of the current best model.
    \item We demonstrate \emph{training time savings} by training a 128 \texttimes\ 128 resolution BigGAN on ImageNet in \nicefrac{1}{4} the time of the baseline, while also achieving superior performance across all image fidelity metrics.
    \item We exhibit the overall computational savings of instance selection by training a 256 \texttimes 256 resolution BigGAN on ImageNet with only 4 V100 GPUs in 11 days. Our model achieves better image fidelity than the baseline model while using \nicefrac{1}{2} as many trainable parameters.
\end{itemize}

\section{Related Work}

Generative modelling of images is a very challenging problem due to the high dimensional nature of images and the complexity of the distributions they form. Several different approaches towards image generation have been proposed, with GANs currently the state-of-the-art in terms of image generation quality. In this work we will focus primarily on GANs, but other types of generative models might also benefit from instance selection prior to model fitting.

\subsection{Sample Filtering in GANs}

One way to improve the sample quality from GANs without making any changes to the architecture or optimization algorithm is by applying techniques which automatically filter out poor quality samples from a trained model.
Discriminator Rejection Sampling (DRS)~\citep{Azadi2019DiscriminatorRS} accomplishes this by performing rejection sampling on the generator. This process is informed by the discriminator, which is reused to estimate density ratios between the real and generated image manifolds. Metropolis-Hastings GAN (MH-GAN)~\citep{Turner2018MetropolisHastingsGA} builds on DRS by i) calibrating the discriminator to achieve more accurate density ratio estimates, and by ii) applying Markov chain Monte Carlo (MCMC) instead of rejection sampling for better performance on high dimensional data. \citet{Ding2020SubsamplingGA} further improve density ratio estimates by fine-tuning a pretrained ImageNet classifier for the task. For more efficient sampling, Discriminator Driven Latent Sampling (DDLS)~\citep{Che2020YourGI} iteratively updates samples in the latent space to push them closer to realistic outputs.

Instead of filtering samples after the GAN has been trained, some methods do so during the training procedure. Latent Optimisation for Generative Adversarial Networks (LOGAN)~\citep{Wu2019LOGANLO} optimizes latent samples each iteration at the cost of an additional forward and backward pass. \citet{Sinha2020TopKTO} demonstrate that gradients from low quality generated samples drive the model away from the nearest mode rather than towards it. As such, gradients from the worst samples each iteration during training may be ignored to improve generation quality.

Perhaps the most well known approach for increasing sample fidelity in GANs is the ``truncation trick''~\citep{brock2018large,karras2019style,marchesi2017megapixel}. The truncation trick is used in the popular models BigGAN~\citep{brock2018large} and StyleGAN~\citep{karras2019style,karras2019analyzing} to improve image quality by manipulating the latent distribution. The original truncation trick as used by BigGAN consists of replacing the latent distribution with a truncated distribution during inference, such that any latent sample that falls outside of some acceptable range is resampled. StyleGAN uses a similar strategy by interpolating samples towards the mean of the latent space instead of resampling them. By moving samples closer to the interior regions of the latent space, sample diversity can effectively be traded for visual fidelity. Our instance selection technique has an effect similar to the truncation trick, but with the added benefit of also reducing model capacity and training time requirements.

\subsection{Instance Selection}
Instance selection is a data preprocessing technique commonly used in the classification setting to select a subset of data from a larger collection~\citep{olvera2010review}. In general, instance selection methods either attempt to reduce the size of the dataset to a more manageable size while retaining informative data points, or try to clean the dataset by eliminating noisy data points. Though commonly used in the setting of big data, instance selection has received little attention from the generative modelling community. \citet{nuha2018training} explore the impact of reducing the size of the training set when training GANs. However, they select data points randomly, and no significant improvement in performance is observed from the removal of data. Core-set selection has been shown to be useful for improving GAN performance when training with small mini-batches, but it ultimately does not improve image fidelity over large mini-batch training \citep{sinha2019small}. Whereas core-set selection attempts to select mini-batches that mimic the distribution of the original dataset, our proposed technique purposefully redefines the target distribution so as to maximize the density of the data manifold.

\section{Instance Selection for GANs}
\label{sec:methodology}

In the context of generative modeling, our motivation is to automatically remove the sparsest regions of the data manifold, specifically those parts that GANs struggle to capture. To do so, we define an image embedding function $F$ and a scoring function $H$.

\textbf{Embedding function $F$} projects images into an embedding space. More precisely, given a dataset of images $\mathcal{X}$, the dataset of embedded images $\mathcal{Z}$ is obtained by applying the embedding function $\mathbf{z}=F(\mathbf{x})$ to each data point $\mathbf{x} \in \mathcal{X}$. For the task of image generation we suggest using perceptually aligned embedding functions~\citep{zhang2018unreasonable}, such as the feature space of a pretrained image classifier.

\textbf{Scoring function $H$}
is used to to assess the manifold density in a neighbourhood around each embedded data point $\mathbf{z}$. In our experiments, we compare three choices of scoring function: log likelihood under a standard Gaussian model, log likelihood under a Probabilistic Principal Component Analysis (PPCA)~\citep{tipping1999probabilistic} model, and distance to the $K$\textsuperscript{th} nearest neighbour (KNN Distance). We select Gaussian and PPCA as simple, well known density models. KNN Distance has previously been used as a measure of local manifold density in classical instance selection~\citep{carbonera2015density}, and has been shown to be useful for defining non-linear image manifolds~\citep{kynkaanniemi2019improved,naeem2020reliable}.

The Gaussian model is fit to the embedded dataset by computing the empirical mean $\boldsymbol{\mu}$ and the sample covariance $\mathbf{\Sigma}$ of $\mathcal{Z}$. The score of each embedded image $\mathbf{z}$ is computed as follows:
\begin{equation}
\label{eq:gaussian_likelihood}
    H_{\text{Gaussian}}(\mathbf{z}) = - \frac{1}{2} [\ln (\vert \mathbf{\Sigma} \vert) + (\mathbf{z} - \boldsymbol{\mu})^{T} \mathbf{\Sigma}^{-1} (\mathbf{z}-\boldsymbol{\mu}) + d \ln(2\pi) ],
\end{equation}
where $d$ is the dimension of $\mathbf{z}$. 

PPCA is fit to the embedded dataset using any standard PPCA solver~\citep{scikit-learn}. We set the number of principal components such that $95\%$ of the variance in the data is preserved. Embedded images are scored as follows:
\begin{equation} 
    H_{\text{PPCA}}(\mathbf{z}) = - \frac{1}{2} [\ln (\vert \mathbf{C} \vert) + \Tr((\mathbf{z} - \boldsymbol{\mu})^{T} \mathbf{C}^{-1} (\mathbf{z}-\boldsymbol{\mu})) + d \ln(2\pi) ], \quad \mathbf{C} = \mathbf{WW}^{T} + \sigma^{2}\mathbf{I},
    \label{eq:ppca_likelihood}
\end{equation}
where $\mathbf{W}$ is the fit model weight matrix, $\boldsymbol{\mu}$ is the empirical mean of $\mathcal{Z}$, $\sigma$ is the residual variance, $\mathbf{I}$ is the identity matrix, and $d$ is the dimension of $\mathbf{z}$.

KNN Distance is used to score data points by calculating the Euclidean distance between $\mathbf{z}$ and $\mathcal{Z} \setminus \{\mathbf{z}\}$, then returning the distance to the $K_{th}$ nearest element. To convert to a score, we make the resulting distance negative, such that smaller distances return larger values. Formally, we can evaluate:\looseness=-1
\begin{equation}
    H_{\text{KNN}}(\mathbf{z}, K, \mathcal{Z}) = -\min_{K} \Big\{ ||\mathbf{z}-\mathbf{z}_{i}||_{2} \quad : \quad \mathbf{z}_{i} \in \mathcal{Z} \setminus \{\mathbf{z}\} \Big\},
    \label{eq:knn_distance}
\end{equation}
where $\min_{K}$ is defined as the $K_{th}$ smallest value in a set. In our experiments we set $K=5$.

To perform \textbf{instance selection}, we compute scores $H(F(\mathbf{x}))$ for each data point and keep all data points with scores above some threshold $\psi$. For convenience, we often set $\psi$ to be equal to some percentile of the scores, such that we preserve the top $N\%$ of the best scoring data points. Thus, given an initial training set consisting of data points $\mathbf{x} \in \mathcal{X}$ we construct our reduced training set $\mathcal{X}'$ by computing: 
\begin{equation}
  \mathcal{X}' = \{ \mathbf{x} \in \mathcal{X} \quad \text{s.t.} \quad H(F(\mathbf{x})) > \psi \}.
\end{equation}

\begin{figure}[t]
     \centering
     \begin{subfigure}[b]{0.48\textwidth}
         \centering
         \includegraphics[width=\textwidth,trim={0 6.9cm 0 0},clip]{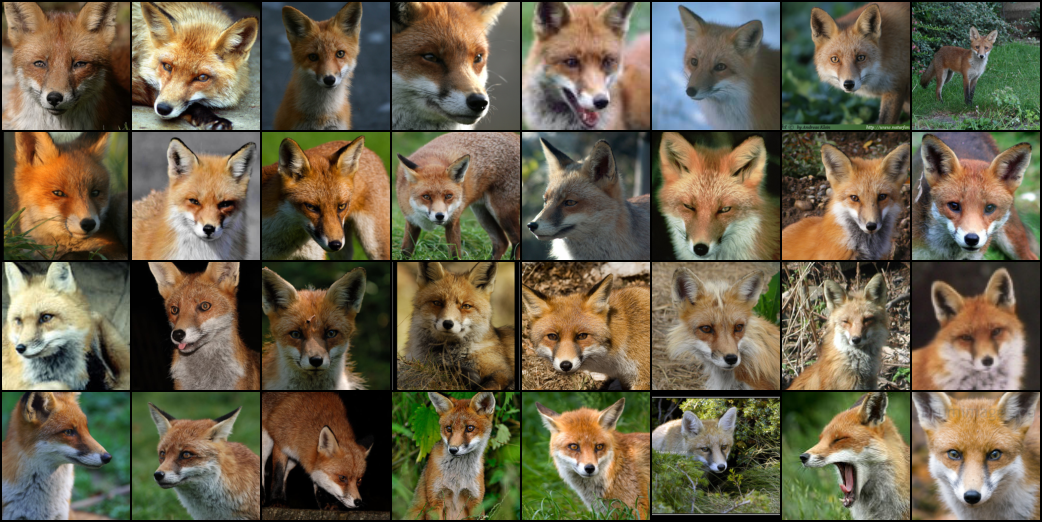}
         \caption{Images with highest likelihood}
         \label{fig:easy_and_hard_example_easy}
     \end{subfigure}
     \quad
     \begin{subfigure}[b]{0.48\textwidth}
         \centering
         \includegraphics[width=\textwidth,trim={0 6.9cm 0 0},clip]{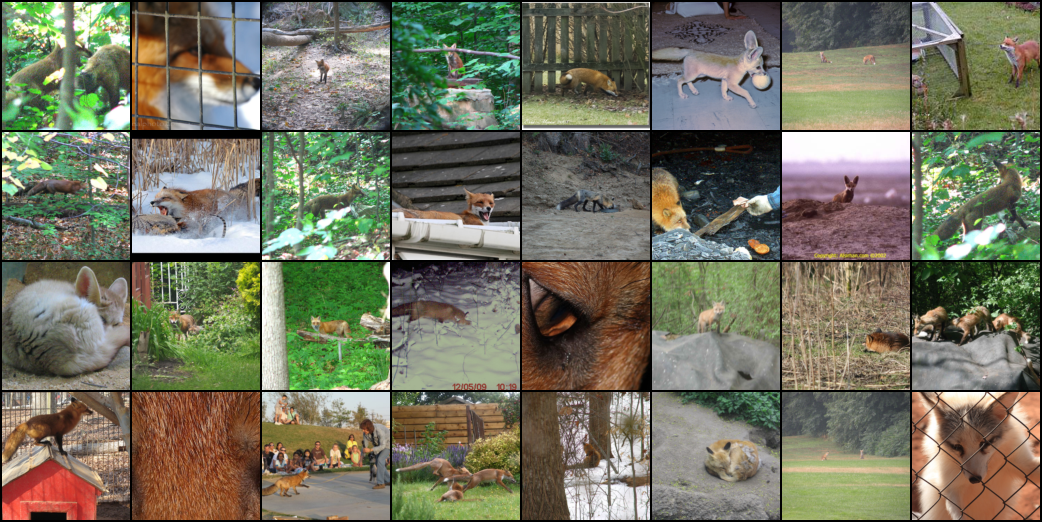}
         \caption{Images with least likelihood}
         \label{fig:easy_and_hard_example_hard}
     \end{subfigure}
        \caption{Examples of the (a) most and (b) least likely resized images of red foxes from the ImageNet dataset, as determined by a Gaussian model fit on images in an Inceptionv3 embedding space. High likelihood images share a similar visual structure, while low likelihood samples are more varied.}
        \label{fig:easy_and_hard_example}
\end{figure}

To illustrate why removing data points from the training set might be a good idea, we look at the most and least likely images from the red fox class of ImageNet (Figure \ref{fig:easy_and_hard_example}). Likelihood is determined by a Gaussian model fit on feature embeddings from a pretrained Inceptionv3 classifier. We notice a stark contrast between the content of the images. The most likely images (\subref{fig:easy_and_hard_example_easy}) are similarly cropped around the fox's face, while the least likely images (\subref{fig:easy_and_hard_example_hard}) have many odd viewpoints and often suffer from occlusion. It is logical to imagine how a generative model trained on these unusual instances may try to generate samples that mimic such conditions, resulting in undesirable outputs.

\section{Experiments}

In this section we review evaluation metrics, motivate selecting instances based on manifold density, and then analyze the impact of applying instance selection to GAN training.

\subsection{Evaluation Metrics}
\label{sec:eval}

We use a variety of evaluation metrics to diagnose the effect that training with instance selection has on the learned distribution, including: (1) Inception Score (IS)~\citep{salimans2016improved}, (2) Fr\'echet Inception Distance (FID)~\citep{heusel2017gans}, (3) Precision and Recall (P\&R)~\citep{kynkaanniemi2019improved}, and (4) Density and Coverage (D\&C)~\citep{naeem2020reliable}. In all cases where a reference distribution is required we use \emph{the original training distribution}. Using the distribution produced after instance selection would unfairly favour the evaluation of instance selection, since the reference distribution could be changed to one that is trivially easy to generate. A detailed description of each evaluation metric is provided in the supplementary material (\S \ref{sec:evlaution_metric_description}).

When calculating FID we follow \citet{brock2018large} in using all images in the training set to estimate the reference distribution, and sampling 50\,k images to make up the generated distribution. For P\&R and D\&C we use an Inceptionv3 embedding.\footnote{We use the PyTorch pretrained Inceptionv3 embedding for all metrics.} $N$ and $M$ are set to 10\,k samples for both the reference and generated distributions, and $K$ is set equal to 5 as recommended by \citet{naeem2020reliable}.

\subsection{Relationship Between Dataset Manifold Density and GAN Performance}
 An image manifold is more accurately defined in regions where many data points are in close proximity to each other~\citep{kynkaanniemi2019improved}. Since GANs attempt to reproduce an image manifold based on data points from a given dataset, we suspect that they should perform better on datasets with well-defined manifolds (i.e.~no sparse manifold regions). To verify this hypothesis, we use the ImageNet\footnote{Use of ImageNet is only for noncommercial, research purposes, and not for training networks deployed in production or for other commercial uses.} 
 dataset~\citep{deng2009imagenet} and treat each of the 1000 classes as a separate dataset. 
 Ideally, we would fit a separate GAN on each class to obtain a ground truth measure of performance, but this is very computationally expensive. Instead, we use a single class-conditional BigGAN from \cite{brock2018large} that has been pretrained on ImageNet at $128\times128$ resolution. For each class, we sample 700 real images from the dataset, and generate 700 class-conditioned samples with the BigGAN. To measure the density for each class manifold we compare three different methods: Gaussian likelihood, Probabilistic Principal Component Analysis (PPCA) likelihood, and distance to the $K$\textsuperscript{th} neighbour (KNN Distance) (\S \ref{sec:methodology}). Images are projected into the feature space of an Inceptionv3 model, and a manifold density score is computed on the features using one of our scoring functions. As an indicator of the true GAN output quality we compute FID between the real and generated distributions for each class.

\begin{figure}[t]
    \centering
    \includegraphics[width=\textwidth]{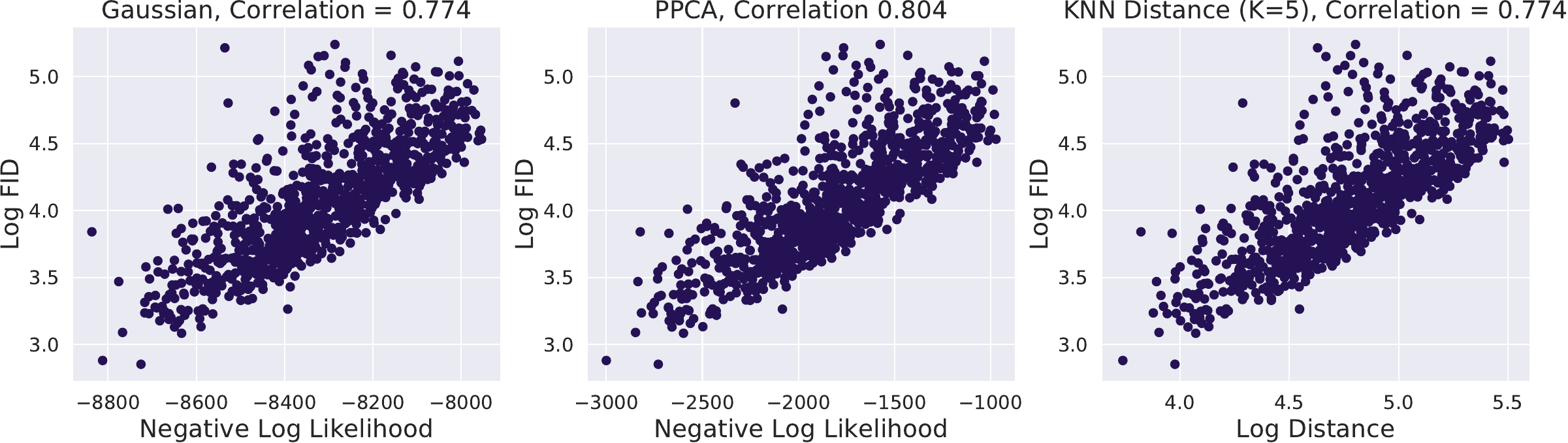}
    \caption{Correlation between manifold density estimates and FID for each class in the ImageNet dataset. Lower values on the x-axis indicate a more dense dataset manifold. Lower values on the y-axis indicate better quality generated samples.}
    \label{fig:fid_density_correlation}
\end{figure}

We observe a strong correlation between each of the manifold density measures and GAN output quality (Figure \ref{fig:fid_density_correlation}). This correlation confirms our hypothesis, suggesting that dataset manifold density is an important factor for achieving high quality generated samples with GANs.

\subsection{Embedding and Scoring Function}
\label{sec:embedding_score_function}

Having established that dataset manifold density is correlated with GAN performance, we explore
artificially increasing the overall density of the training set by removing data points that lie in low density regions of the data manifold. To this end, we train several Self-Attention GANs (SAGAN)~\citep{zhang2018self} on ImageNet at 64 \texttimes\ 64 resolution. Each model is trained on a different $50\%$ subset of ImageNet, as chosen by instance selection using different embedding and scoring functions as described in \S \ref{sec:methodology}. Instance selection is applied per-class. We use the default settings for SAGAN, except that we use a batch size of 128 instead of 256, apply the self-attention module at 32 \texttimes\ 32 resolution instead of 64 \texttimes\ 64, and reduce the number of channels in each layer by half in order to reduce the computational cost of our initial exploratory experiments. All models are trained for 200k iterations.
The results of these experiments are shown in Table \ref{table:embed_scoring_sweep}. For reference, we include scores achieved by real (i.e.~not generated) data in Table \ref{table:real_data} in the supplementary material.

\begin{table}[b]
\centering
\caption{Comparison of embedding and scoring functions on 64 \texttimes\ 64 ImageNet image generation task. All tests train a SAGAN model for 200k iterations. Models trained with instance selection significantly outperform models trained without instance selection, despite training on a fraction of the available data. RR is the retention ratio (percentage of dataset trained on). 
Best results in bold.}
\label{table:embed_scoring_sweep}
\begin{tabular}{lccc|c|c|cc|cc}
\toprule
\begin{tabular}[c]{@{}l@{}}Instance \\ Selection\end{tabular} & \begin{tabular}[c]{@{}c@{} }RR \\ (\%)\end{tabular} & Embedding & Pretraining & IS $\uparrow$ & FID $\downarrow$ & P $\uparrow$ & R $\uparrow$ & D $\uparrow$ & C $\uparrow$\\
\midrule
None & 100 & - & -  & 15.4 & 21.4 & 0.66 & \textbf{0.62} & 0.64 & 0.64 \\ 
Uniform & 50 & - & - & 15.5 & 22.8 & 0.65 & \textbf{0.62} & 0.65 & 0.65 \\ \midrule
Gaussian & 50 & Inceptionv3 & ImageNet & \textbf{25.7} & \textbf{12.6} & \textbf{0.77} & 0.59 & \textbf{0.97} & \textbf{0.83} \\
PPCA & 50 & Inceptionv3 & ImageNet  & 25.5 & 13.2 & 0.76 & 0.58 & \textbf{0.97} & 0.82 \\
KNN Dist & 50 & Inceptionv3 & ImageNet & 25.4 & 13.1 & 0.76 & 0.58 & \textbf{0.97} & 0.82 \\
\midrule
Gaussian & 50 & Inceptionv3 & Random init & 15.5 & 21.9 & 0.66 & 0.61 & 0.68 & 0.65 \\
Gaussian  & 50 & ResNet-50 & Places365 & 20.6 & 16.5 & 0.74 & 0.59 & 0.88 & 0.76 \\
Gaussian  & 50 & ResNet-50 & SwAV & 20.3 & 16.7 & 0.74 & 0.57 & 0.89 & 0.76 \\
Gaussian  & 50 & ResNet-50 & ImageNet & 22.0 & 14.6 & 0.76 & 0.59 & 0.92 & 0.79 \\
Gaussian & 50 & ResNeXt-101 & Instagram 1B & 24.1 & 14.1 & 0.73 & 0.61 & 0.86 & 0.80 \\  
\bottomrule
\end{tabular}
\end{table}

All runs utilizing instance selection significantly outperform the baseline model trained on the full dataset, despite only having access to half as much training data (Table \ref{table:embed_scoring_sweep}). We observe a large increase in image fidelity, as indicated by the improvements in Inception Score, Precision, and Density, and a slight drop in overall diversity, as measured by Recall. Coverage, which measures realism-constrained diversity, benefits greatly from the more realistic samples and thus sees an increase, despite the reduction in overall diversity. Since the increase in image quality is much greater than the decrease in diversity, FID also improves. To verify that the gains are not simply caused by the reduction in dataset size we train a model on a 50\% subset that was uniform-randomly sampled from the full dataset. Here, we observe little change in performance compared to the baseline, indicating that performance improvements are indeed due to careful selection of training data, rather than the reduction of dataset size.

We find that all three candidate scoring functions: Gaussian likelihood, PPCA likelihood, and KNN distance, significantly outperform the full dataset baseline. Gaussian likelihood slightly outperforms the alternatives, so we use it as the scoring function in the remainder of our experiments.

To understand the importance of the embedding function, we compare several different model embeddings that have been trained on different datasets: Inceptionv3~\citep{szegedy2016rethinking} trained on ImageNet, ResNet50~\citep{he2016deep} trained on Places365~\citep{zhou2017places},
ImageNet, and with SwAV unsupervised pretraining~\citep{caron2020unsupervised}, 
and ResNeXt-101 32x8d~\citep{xie2017aggregated} trained with weak supervision on Instagram 1B~\citep{mahajan2018exploring}. We also compare a randomly initialized Inceptionv3 with no pretraining as a random embedding. For all architectures, features are extracted after the global average pooling layer. We find that all feature embeddings improve performance over the full dataset baseline except for the randomly initialized network. These results suggest that an embedding function that is well aligned with the target domain is required in order for instance selection to be effective. The ImageNet pretrained Inceptionv3 embedding performs best overall, and was chosen as the embedding function for the rest of our experiments. We note that using an Inceptionv3 embedding both in instance selection and in the evaluation metrics may yield some non-negligible advantage in evaluation, since selected instances are those that the network prefers.

\subsection{Retention Ratio}
\label{sec:retention_rate}

An important consideration when performing instance selection is determining what proportion of the original dataset to keep, a hyperparameter which we call retention ratio. To investigate the impact of the retention ratio on training, we train ten SAGANs on ImageNet, each retaining different amounts of the original dataset in $10\%$ intervals. GAN hyperparameters are the same as in \S \ref{sec:embedding_score_function}, except that we extend training until 500k iterations in order to observe model behaviours over a longer training window. Results are shown in Figure \ref{fig:keep_percentage_plot} and Table \ref{table:retention_ratio_sweep_results} in the supplementary material.

\begin{figure}[!b]
    \centering
    \includegraphics[width=0.97\textwidth]{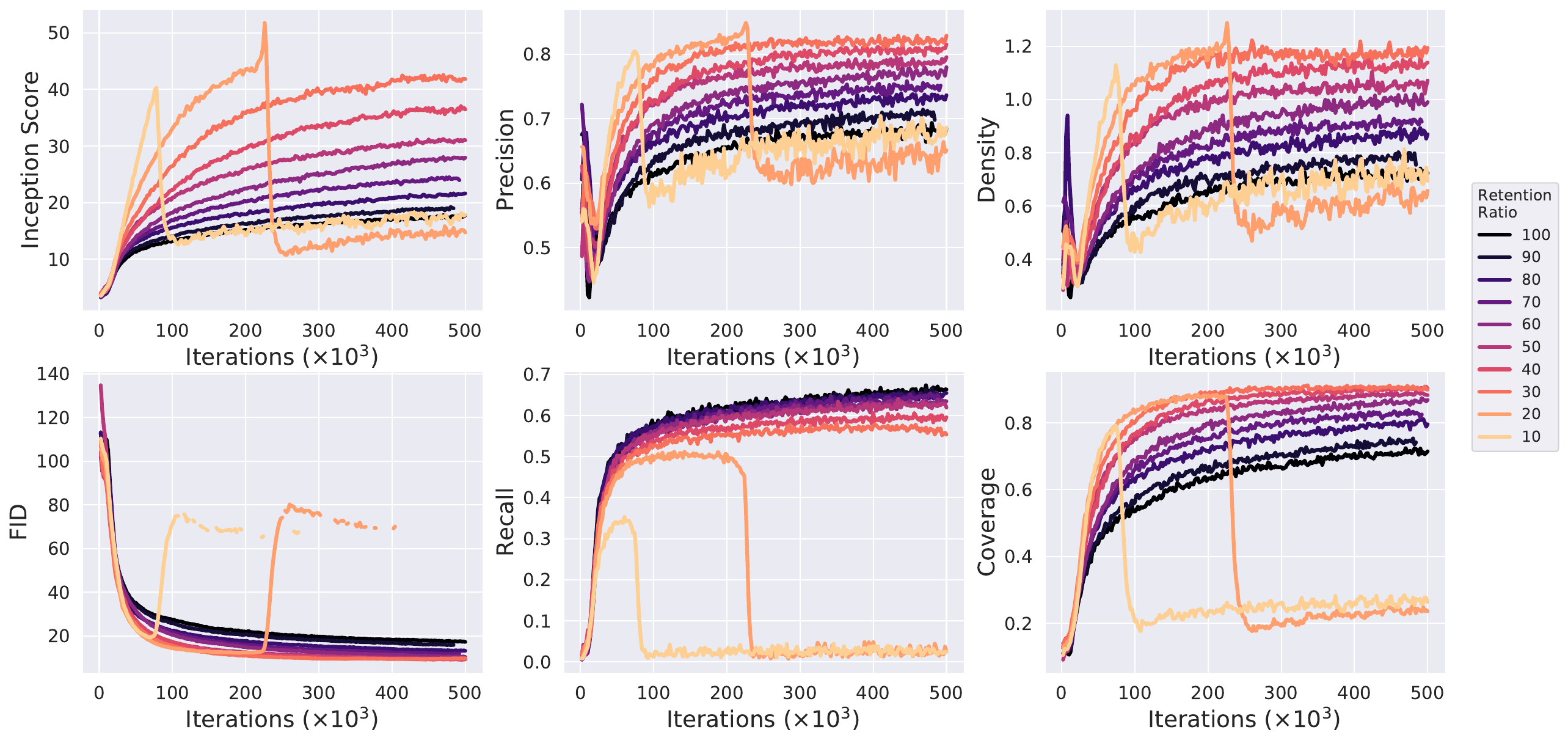}
    \caption{SAGAN trained on $64\times64$ ImageNet, with instance selection used to reduced the dataset by varying amounts. Retention ratio = 100 indicates a model trained on the full dataset (i.e.~no instance selection). The application of instance selection boosts overall performance significantly.}
    \label{fig:keep_percentage_plot}
\end{figure}

As larger portions of the original dataset are removed we see consistent improvements in image fidelity (increasing Inception Score, Precision, and Density) and reductions in sample diversity (decreasing Recall). Interestingly, metrics which take into account both realism and diversity (FID and Coverage) continue to see gains until roughly $70\%$ of the dataset has been removed, at which point they begin to decrease. This behaviour suggests that, given the ability of current state-of-the-art models to learn from limited data, sample fidelity is valued much more than diversity. When too much of the dataset is removed some models collapse prematurely, likely due to the discriminator quickly overfitting the small training set. It is expected that applying data augmentation could resolve this issue~\citep{karras2020training,zhao2020differentiable}. To further improve image fidelity, instance selection could be combined with the truncation trick (\S \ref{sec:truncation_analysis}).

\begin{figure}[t]
     \centering
     \begin{subfigure}[b]{0.45\textwidth}
         \centering
         \includegraphics[width=\textwidth]{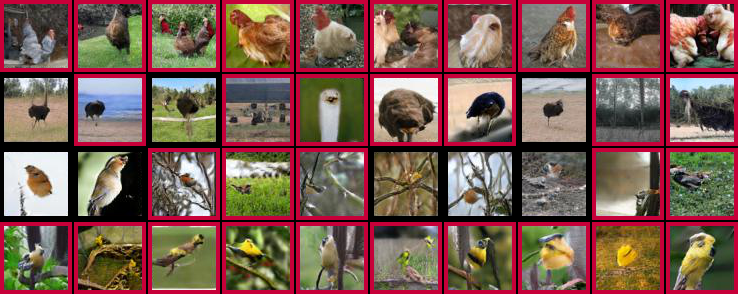}
         \caption{Baseline (trained on $100\%$ of dataset)}
     \end{subfigure}
     \quad
     \begin{subfigure}[b]{0.45\textwidth}
         \centering
         \includegraphics[width=\textwidth]{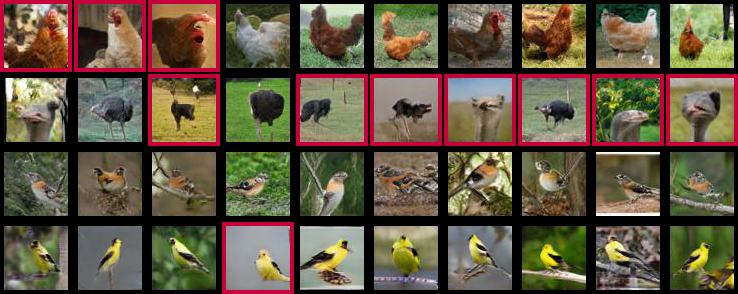}
         \caption{Instance Selection (trained on $40\%$ of dataset)}
     \end{subfigure}
        \caption{Samples of bird classes from SAGAN trained on $64\times64$ ImageNet. Each row is conditioned on a different class. Red borders indicate misclassification by a row-specific pretrained Inceptionv3 classifier. Instance selection (b) significantly improves sample fidelity and class consistency compared to the baseline (a).}
        \label{fig:imagenet_64_sample_comparison}
\end{figure}

Our best performing SAGAN model in terms of FID was trained on only $40\%$ of the ImageNet dataset, yet \emph{outperforms} FQ-BigGAN~\citep{zhao2020feature}, the current state-of-the-art model for the task of $64\times64$ ImageNet generation. Despite using $2\times$ less parameters and a $4\times$ smaller batch size, our SAGAN achieves a better FID (9.07 vs.~9.76). As indicated by these scores and the errors made by a pretrained classifier, samples from our instance selection model are significantly more recognizable than those from the baseline model trained on the full dataset (Figure \ref{fig:imagenet_64_sample_comparison}).\looseness=-1

\subsection{128 \texttimes\ 128 ImageNet}
To examine the impact of instance selection on the training time of large-scale models, we train two BigGAN models on 128 \texttimes\ 128 ImageNet\footnote{We use the official BigGAN implementation from \url{https://github.com/ajbrock/BigGAN-PyTorch}.}. Our baseline model uses the default hyperparameters from BigGAN~\citep{brock2018large}, with the exception that we reduce the channel multiplier from 96 to 64 (i.e.~half of the capacity) and only use a single discriminator update instead of two for faster training. Our instance selection model uses the same settings as the baseline, but is trained on $50\%$ of the dataset. Although large batch sizes are critical for achieving good performance with the baseline BigGAN~\cite{brock2018large}, we found them to degrade performance when combined with instance selection. Therefore, we reduce the batch size from BigGAN's default of 2048 to 256 for the instance selection model. Both models are trained on 8 NVIDIA V100 GPUs with 16GB of RAM, using gradient accumulation to achieve the necessary batch sizes.

Despite using a much smaller batch size, our model trained with instance selection outperforms the baseline in all metrics except for Recall (Table \ref{table:128x128_imagenet}), as expected due to the diversity/fidelity trade-off. The instance selection model trains significantly faster than the baseline, requiring less than four days while the baseline requires more than two weeks. 

\begin{table}[h]
\centering
\caption{Performance of models on the 128 \texttimes\ 128 ImageNet image generation task. Both models use a channel multiplier of 64 and a single discriminator update per generator update. The baseline model uses a batch size of 2048, while the instance selection model uses a batch size of 256.}
\label{table:128x128_imagenet}
\begin{tabular}{l|c|c|cc|cc|cc}
\toprule
Model & IS $\uparrow$  & FID $\downarrow$ & P $\uparrow$  & R $\uparrow$  & D $\uparrow$  & C $\uparrow$  & Time $\downarrow$ & Hardware \\
\midrule
BigGAN & 68.8 & 11.5 & 0.76 &\textbf{ 0.66} & 0.9 & 0.84 & 14.8 days & 8 V100 \\
BigGAN + Inst.~Sel. & \textbf{114.3} & \textbf{9.6} & \textbf{0.88} & 0.50 & \textbf{1.34} & \textbf{0.90} & \textbf{3.7 days} & 8 V100 \\
\bottomrule
\end{tabular}
\end{table}

\subsection{256 \texttimes 256 ImageNet}
\label{sec:256x256_imagenet}

To further demonstrate instance selection we train a BigGAN on ImageNet at 256 \texttimes\ 256 resolution using 4 V100s with 32GB of RAM each. Since training a baseline model without instance selection with the same hardware setup would take an excessively long time (1-2 months), we instead compare to the 256 \texttimes\ 256 BigGAN from \citet{brock2018large} using the official pretrained weights\footnote{Pretrained BigGAN weights from \url{https://colab.research.google.com/github/tensorflow/hub/blob/master/examples/colab/biggan_generation_with_tf_hub.ipynb}}. Compared to this baseline, our model uses half the capacity (channel multiplier reduced from 96 to 64), 8\texttimes\ smaller batch size (from 2048 to 256), and applies the self-attention block in the generator at a resolution of 64 \texttimes\ 64 instead of 128 \texttimes\ 128. The retention ratio for instance selection is set to $50\%$. Similar to the baseline, we use  two discriminator update steps per generator update for this experiment. Quantitative results are presented in Table \ref{table:256x256_imagenet}, and samples are shown in Figure \ref{fig:256x256_samples} and \S \ref{sec:sample_sheets} in the supplementary material.

Our instance selection model trains in less than 11 days, and uses approximately one order of magnitude less multiply-accumulate operations (MACS) than the baseline throughout the duration of training. Despite having half as much capacity, our model outperforms the baseline in all image fidelity focused metrics (Inception Score, Precision, and Density), and achieves comparable performance on metrics that jointly consider image quality and diversity (FID and Coverage). As expected, the better image quality comes at the cost of overall sample diversity (indicated by Recall). To our knowledge, this is the first time photorealistic generation of 256 \texttimes\ 256 ImageNet images has been achieved without the use of specialized hardware (i.e. hundreds of TPUs).

\begin{figure}[t]
     \centering
     \begin{subfigure}[b]{0.48\textwidth}
         \centering
         \includegraphics[width=\textwidth]{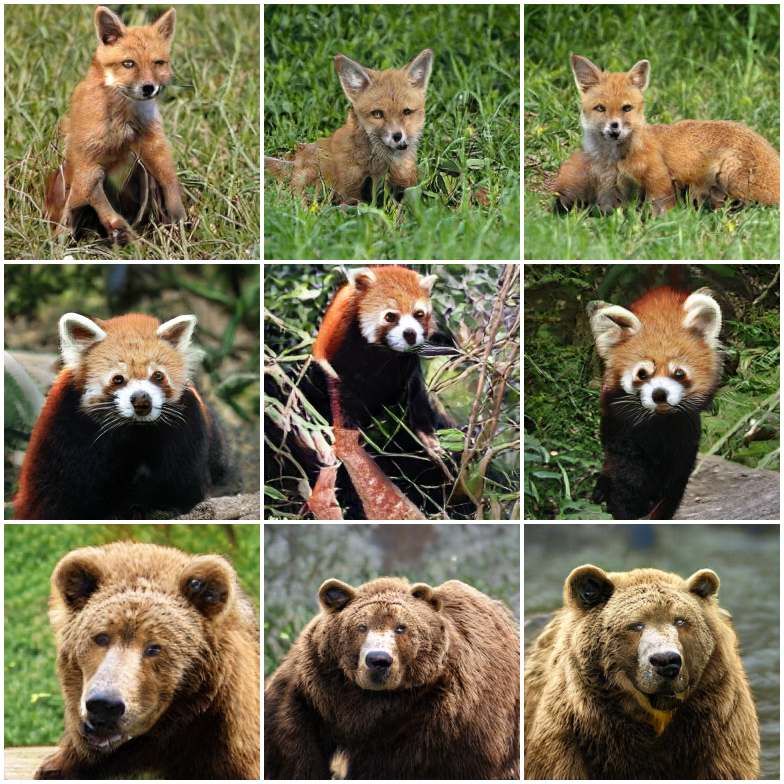}
         \caption{Baseline (trained on $100\%$ of dataset)}
     \end{subfigure}
     \quad
     \begin{subfigure}[b]{0.48\textwidth}
         \centering
         \includegraphics[width=\textwidth]{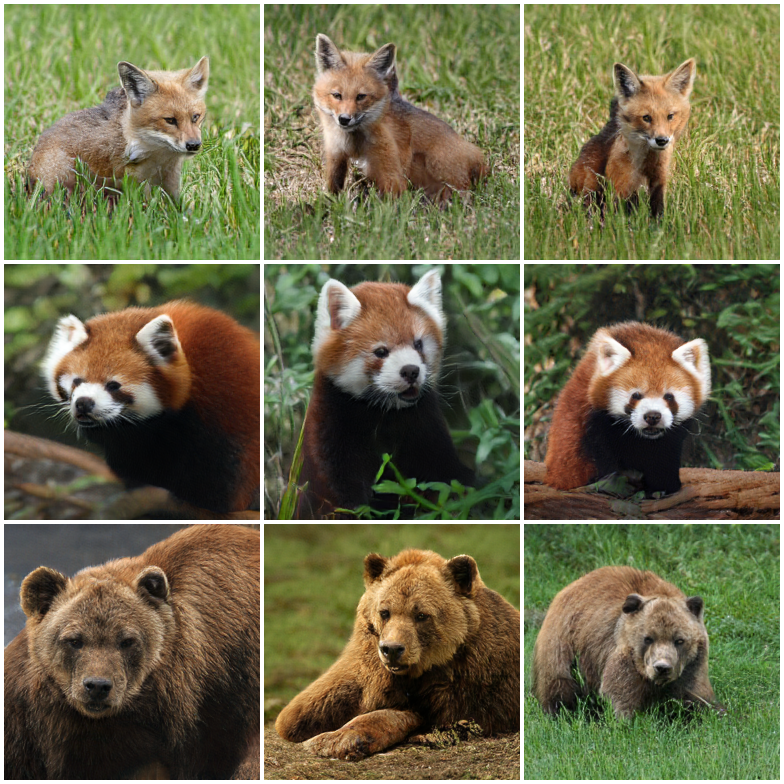}
         \caption{Instance Selection (trained on $50\%$ of dataset)}
     \end{subfigure}
        \caption{Samples from BigGAN trained on 256 \texttimes\ 256 ImageNet, with the truncation trick. Samples are selected to demonstrate the highest quality outputs for each model. The baseline model (a) struggles to produce convincing facial details, which the instance selection model (b) successfully achieves. Zoom in for best viewing.}
        \label{fig:256x256_samples}
\end{figure}

\begin{table}[h]
\centering
\caption{Performance of models for 256 \texttimes\ 256 ImageNet image generation. The instance selection model uses half as many parameters as the baseline model. All metrics are computed using PyTorch Inceptionv3 embeddings, and may therefore differ from numbers computed with TensorFlow.} 
\label{table:256x256_imagenet}
\begin{tabular}{l|c|c|cc|cc|cc}
\toprule
Model & IS $\uparrow$  & FID $\downarrow$ & P $\uparrow$  & R $\uparrow$  & D $\uparrow$  & C $\uparrow$  & Time & Hardware \\
\midrule
BigGAN & 135.4 & \textbf{9.8} & 0.86 &\textbf{0.70} & 1.18 & 0.92 & 1-2 days & 256 TPUv3 \\
BigGAN + Inst.~Sel. & \textbf{165.3} & 10.6 & \textbf{0.91} & 0.52 & \textbf{1.48} & \textbf{0.93} & 10.7 days & 4 V100 \\
\bottomrule
\end{tabular}
\end{table}

\section{Instance Selection in Practice}
As the experiments have shown, instance selection stands as a useful tool for trading away sample diversity in exchange for improvements in image fidelity, faster training, and lower model capacity requirements. We believe that this trade-off is a worthwhile hyperparameter to tune in consideration of the available compute budget, just as it is common practice to adjust model capacity or batch size to fit within the memory constraints of the available hardware. 

The control over the diversity/fidelity trade-off afforded by instance selection also yields a tool that can be used to better understand the behaviour and limitations of existing evaluation metrics. For instance, in some cases when applying instance selection, we observed that certain diversity-sensitive metrics (such as FID and Coverage) improved, even though the diversity of the training set had been significantly reduced. We leave it for future work to determine whether this is a limitation of these metrics, or a behaviour that should be expected. 

Finally, instance selection can be used to automatically curate new datasets for the task of image generation. Existing datasets that are designed for image synthesis often use manual filtering and hand-crafted cropping and alignment tools to increase the dataset manifold density~\cite{karras2019style}. As an alternative to these time-intensive procedures, instance selection provides a generic solution that can quickly be applied to any uncurated set of images.

\section{Conclusion}

Folk wisdom suggests \emph{more data is better}, however, it is known that areas of the data manifold that are sparsely represented pose a challenge to current GANs \citep{karras2019style}. To directly address this challenge we introduce a new tool: dataset curation via instance selection. Our motivation is to remove sparse regions of the data manifold before training, acknowledging that they will ultimately be poorly represented by the GAN, and therefore, that attempting to capture them is an inefficient use of model capacity. Moreover, popular post-processing methods such as rejection sampling or latent space truncation will likely ignore these regions as represented by the model. There are multiple benefits of taking the instance selection approach: (1) We improve sample fidelity across a variety of metrics compared to training on uncurated data; (2) We demonstrate that reallocating model capacity to denser regions of the data manifold leads to efficiency gains, meaning that we can achieve SOTA quality with smaller-capacity models trained in far less time. To our knowledge, instance selection has not yet been formally analyzed in the generative setting. However, we argue that it is more important here than in supervised learning because of the absence of an annotation phase where humans often perform some kind of formal or informal curation.

We have only considered the setting where curation is performed up-front, prior to training. However, our results suggest that dynamic curation, including curriculum learning informed by the kinds of perceptually aligned embeddings we consider here, is an interesting direction for future work.

\section*{Broader Impact}
The application of instance selection to the task of image generative modelling brings with it several benefits. Gains in image generation quality are an obvious improvement, but perhaps more impactful to the broader community are the reductions in model capacity and training time that are afforded. Reducing the computational barrier to entry for training large-scale generative models provides many individuals, including students, AI artists, and ML enthusiasts, with access to models that are otherwise restricted to only the most well resourced labs. In addition to greater accessibility, lowering the computational requirements for training large-scale generative models also reduces associated energy costs and \ce{CO2} emissions associated with the training process. 

One side effect of our instance selection method is that, by nature of design, generated results are more likely to reflect the content that makes up the majority of the training set. As such, dataset bias is amplified as instances that are poorly represented in the dataset may be completely ignored. However, this limitation can be addressed by properly balancing the training set before instance selection is applied or alternatively, ensuring a more diverse \& inclusive data collection effort to begin with.

As with any form of generative model, there is some potential for misuse. A common example is ``deepfakes'', where a generative model is used to manipulate images or videos well enough that humans cannot distinguish real from fake. While often used to create humorous videos in which actors' faces are swapped, deepfakes also have the potential for more nefarious uses, such as for blackmail or spreading misinformation. Fortunately, much recent effort has been dedicated towards automatic detection of these false images \citep{tolosana2020deepfakes}. These techniques attempt to find manipulated media by detecting inconsistencies, such as in the synchronization of lip movement and speech audio, or generation artifacts, such as missing reflections or other minute details.

\begin{ack}
The authors would like to thank Colin Brennan, with whom discussions about dataset learnability kicked off this project, and Brendan Duke, for being a constant sounding board. Resources used in preparing this research were provided to GWT and TD, in part, by NSERC, the Canada Foundation for Innovation, the Province of Ontario, the Government of Canada through CIFAR, Compute Canada, and companies sponsoring the Vector Institute: \url{http://www.vectorinstitute.ai/\#partners}.
\end{ack}

{\small
\bibliographystyle{plainnat}
\bibliography{biblio}

\begin{thebibliography}{41}
\providecommand{\natexlab}[1]{#1}
\providecommand{\url}[1]{\texttt{#1}}
\expandafter\ifx\csname urlstyle\endcsname\relax
  \providecommand{\doi}[1]{doi: #1}\else
  \providecommand{\doi}{doi: \begingroup \urlstyle{rm}\Url}\fi

\bibitem[Azadi et~al.(2019)Azadi, Olsson, Darrell, Goodfellow, and
  Odena]{Azadi2019DiscriminatorRS}
Samaneh Azadi, Catherine Olsson, Trevor Darrell, Ian~J. Goodfellow, and
  Augustus Odena.
\newblock Discriminator rejection sampling.
\newblock \emph{ArXiv}, abs/1810.06758, 2019.

\bibitem[Brock et~al.(2018)Brock, Donahue, and Simonyan]{brock2018large}
Andrew Brock, Jeff Donahue, and Karen Simonyan.
\newblock Large scale {GAN} training for high fidelity natural image synthesis.
\newblock \emph{ICLR}, 2018.

\bibitem[Carbonera and Abel(2015)]{carbonera2015density}
Joel~Luis Carbonera and Mara Abel.
\newblock A density-based approach for instance selection.
\newblock In \emph{2015 IEEE 27th International Conference on Tools with
  Artificial Intelligence (ICTAI)}, pages 768--774. IEEE, 2015.

\bibitem[Caron et~al.(2020)Caron, Misra, Mairal, Goyal, Bojanowski, and
  Joulin]{caron2020unsupervised}
Mathilde Caron, Ishan Misra, Julien Mairal, Priya Goyal, Piotr Bojanowski, and
  Armand Joulin.
\newblock Unsupervised learning of visual features by contrasting cluster
  assignments.
\newblock \emph{arXiv preprint arXiv:2006.09882}, 2020.

\bibitem[Che et~al.(2020)Che, Zhang, Sohl-Dickstein, Larochelle, Paull, Cao,
  and Bengio]{Che2020YourGI}
Tong Che, Ruixiang Zhang, Jascha Sohl-Dickstein, Hugo Larochelle, Liam Paull,
  Yuan Cao, and Yoshua Bengio.
\newblock Your gan is secretly an energy-based model and you should use
  discriminator driven latent sampling.
\newblock \emph{ArXiv}, abs/2003.06060, 2020.

\bibitem[Clark et~al.(2019)Clark, Donahue, and Simonyan]{clark2019efficient}
Aidan Clark, Jeff Donahue, and Karen Simonyan.
\newblock Efficient video generation on complex datasets.
\newblock \emph{arXiv preprint arXiv:1907.06571}, 2019.

\bibitem[Deng et~al.(2009)Deng, Dong, Socher, Li, Li, and
  Fei-Fei]{deng2009imagenet}
Jia Deng, Wei Dong, Richard Socher, Li-Jia Li, Kai Li, and Li~Fei-Fei.
\newblock Imagenet: A large-scale hierarchical image database.
\newblock In \emph{2009 IEEE conference on computer vision and pattern
  recognition}, pages 248--255. Ieee, 2009.

\bibitem[Ding et~al.(2020)Ding, Wang, and Welch]{Ding2020SubsamplingGA}
Xin Ding, Z.~Jane Wang, and William~J. Welch.
\newblock Subsampling generative adversarial networks: Density ratio estimation
  in feature space with softplus loss.
\newblock \emph{IEEE Transactions on Signal Processing}, 68:\penalty0
  1910--1922, 2020.

\bibitem[He et~al.(2016)He, Zhang, Ren, and Sun]{he2016deep}
Kaiming He, Xiangyu Zhang, Shaoqing Ren, and Jian Sun.
\newblock Deep residual learning for image recognition.
\newblock In \emph{Proceedings of the IEEE conference on computer vision and
  pattern recognition}, pages 770--778, 2016.

\bibitem[Heusel et~al.(2017)Heusel, Ramsauer, Unterthiner, Nessler, and
  Hochreiter]{heusel2017gans}
Martin Heusel, Hubert Ramsauer, Thomas Unterthiner, Bernhard Nessler, and Sepp
  Hochreiter.
\newblock Gans trained by a two time-scale update rule converge to a local nash
  equilibrium.
\newblock In \emph{Advances in neural information processing systems}, pages
  6626--6637, 2017.

\bibitem[Karras et~al.(2019{\natexlab{a}})Karras, Laine, and
  Aila]{karras2019style}
Tero Karras, Samuli Laine, and Timo Aila.
\newblock A style-based generator architecture for generative adversarial
  networks.
\newblock In \emph{Proceedings of the IEEE Conference on Computer Vision and
  Pattern Recognition}, pages 4401--4410, 2019{\natexlab{a}}.

\bibitem[Karras et~al.(2019{\natexlab{b}})Karras, Laine, Aittala, Hellsten,
  Lehtinen, and Aila]{karras2019analyzing}
Tero Karras, Samuli Laine, Miika Aittala, Janne Hellsten, Jaakko Lehtinen, and
  Timo Aila.
\newblock Analyzing and improving the image quality of stylegan.
\newblock \emph{arXiv preprint arXiv:1912.04958}, 2019{\natexlab{b}}.

\bibitem[Karras et~al.(2020)Karras, Aittala, Hellsten, Laine, Lehtinen, and
  Aila]{karras2020training}
Tero Karras, Miika Aittala, Janne Hellsten, Samuli Laine, Jaakko Lehtinen, and
  Timo Aila.
\newblock Training generative adversarial networks with limited data.
\newblock \emph{arXiv preprint arXiv:2006.06676}, 2020.

\bibitem[Kynk{\"a}{\"a}nniemi et~al.(2019)Kynk{\"a}{\"a}nniemi, Karras, Laine,
  Lehtinen, and Aila]{kynkaanniemi2019improved}
Tuomas Kynk{\"a}{\"a}nniemi, Tero Karras, Samuli Laine, Jaakko Lehtinen, and
  Timo Aila.
\newblock Improved precision and recall metric for assessing generative models.
\newblock In \emph{Advances in Neural Information Processing Systems}, pages
  3929--3938, 2019.

\bibitem[Mahajan et~al.(2018)Mahajan, Girshick, Ramanathan, He, Paluri, Li,
  Bharambe, and van~der Maaten]{mahajan2018exploring}
Dhruv Mahajan, Ross Girshick, Vignesh Ramanathan, Kaiming He, Manohar Paluri,
  Yixuan Li, Ashwin Bharambe, and Laurens van~der Maaten.
\newblock Exploring the limits of weakly supervised pretraining.
\newblock In \emph{Proceedings of the European Conference on Computer Vision
  (ECCV)}, pages 181--196, 2018.

\bibitem[Marchesi(2017)]{marchesi2017megapixel}
Marco Marchesi.
\newblock Megapixel size image creation using generative adversarial networks.
\newblock \emph{arXiv preprint arXiv:1706.00082}, 2017.

\bibitem[McInnes et~al.(2018)McInnes, Healy, and Melville]{mcinnes2018umap}
Leland McInnes, John Healy, and James Melville.
\newblock Umap: Uniform manifold approximation and projection for dimension
  reduction.
\newblock \emph{arXiv preprint arXiv:1802.03426}, 2018.

\bibitem[Mescheder et~al.(2018)Mescheder, Geiger, and
  Nowozin]{mescheder2018training}
Lars Mescheder, Andreas Geiger, and Sebastian Nowozin.
\newblock Which training methods for gans do actually converge?
\newblock \emph{arXiv preprint arXiv:1801.04406}, 2018.

\bibitem[Naeem et~al.(2020)Naeem, Oh, Uh, Choi, and Yoo]{naeem2020reliable}
Muhammad~Ferjad Naeem, Seong~Joon Oh, Youngjung Uh, Yunjey Choi, and Jaejun
  Yoo.
\newblock Reliable fidelity and diversity metrics for generative models.
\newblock \emph{arXiv preprint arXiv:2002.09797}, 2020.

\bibitem[Nuha et~al.(2018)]{nuha2018training}
Fajar~Ulin Nuha et~al.
\newblock Training dataset reduction on generative adversarial network.
\newblock \emph{Procedia computer science}, 144:\penalty0 133--139, 2018.

\bibitem[Olvera-L{\'o}pez et~al.(2010)Olvera-L{\'o}pez, Carrasco-Ochoa,
  Mart{\'\i}nez-Trinidad, and Kittler]{olvera2010review}
J~Arturo Olvera-L{\'o}pez, J~Ariel Carrasco-Ochoa, J~Francisco
  Mart{\'\i}nez-Trinidad, and Josef Kittler.
\newblock A review of instance selection methods.
\newblock \emph{Artificial Intelligence Review}, 34\penalty0 (2):\penalty0
  133--143, 2010.

\bibitem[Pedregosa et~al.(2011)Pedregosa, Varoquaux, Gramfort, Michel, Thirion,
  Grisel, Blondel, Prettenhofer, Weiss, Dubourg, Vanderplas, Passos,
  Cournapeau, Brucher, Perrot, and Duchesnay]{scikit-learn}
F.~Pedregosa, G.~Varoquaux, A.~Gramfort, V.~Michel, B.~Thirion, O.~Grisel,
  M.~Blondel, P.~Prettenhofer, R.~Weiss, V.~Dubourg, J.~Vanderplas, A.~Passos,
  D.~Cournapeau, M.~Brucher, M.~Perrot, and E.~Duchesnay.
\newblock Scikit-learn: Machine learning in {P}ython.
\newblock \emph{Journal of Machine Learning Research}, 12:\penalty0 2825--2830,
  2011.

\bibitem[Ravuri and Vinyals(2019)]{ravuri2019classification}
Suman Ravuri and Oriol Vinyals.
\newblock Classification accuracy score for conditional generative models.
\newblock In \emph{Advances in Neural Information Processing Systems}, pages
  12268--12279, 2019.

\bibitem[Salimans et~al.(2016)Salimans, Goodfellow, Zaremba, Cheung, Radford,
  and Chen]{salimans2016improved}
Tim Salimans, Ian Goodfellow, Wojciech Zaremba, Vicki Cheung, Alec Radford, and
  Xi~Chen.
\newblock Improved techniques for training gans.
\newblock In \emph{Advances in neural information processing systems}, pages
  2234--2242, 2016.

\bibitem[Shmelkov et~al.(2018)Shmelkov, Schmid, and Alahari]{shmelkov2018good}
Konstantin Shmelkov, Cordelia Schmid, and Karteek Alahari.
\newblock How good is my gan?
\newblock In \emph{Proceedings of the European Conference on Computer Vision
  (ECCV)}, pages 213--229, 2018.

\bibitem[Sinha et~al.(2019)Sinha, Zhang, Goyal, Bengio, Larochelle, and
  Odena]{sinha2019small}
Samarth Sinha, Han Zhang, Anirudh Goyal, Yoshua Bengio, Hugo Larochelle, and
  Augustus Odena.
\newblock Small-gan: Speeding up gan training using core-sets.
\newblock \emph{arXiv preprint arXiv:1910.13540}, 2019.

\bibitem[Sinha et~al.(2020)Sinha, Goyal, Raffel, and Odena]{Sinha2020TopKTO}
Samarth Sinha, Anirudh Goyal, Colin Raffel, and Augustus Odena.
\newblock Top-k training of gans: Improving generators by making critics less
  critical.
\newblock \emph{ArXiv}, abs/2002.06224, 2020.

\bibitem[Szegedy et~al.(2016)Szegedy, Vanhoucke, Ioffe, Shlens, and
  Wojna]{szegedy2016rethinking}
Christian Szegedy, Vincent Vanhoucke, Sergey Ioffe, Jon Shlens, and Zbigniew
  Wojna.
\newblock Rethinking the inception architecture for computer vision.
\newblock In \emph{Proceedings of the IEEE conference on computer vision and
  pattern recognition}, pages 2818--2826, 2016.

\bibitem[Tipping and Bishop(1999)]{tipping1999probabilistic}
Michael~E Tipping and Christopher~M Bishop.
\newblock Probabilistic principal component analysis.
\newblock \emph{Journal of the Royal Statistical Society: Series B (Statistical
  Methodology)}, 61\penalty0 (3):\penalty0 611--622, 1999.

\bibitem[Tolosana et~al.(2020)Tolosana, Vera-Rodriguez, Fierrez, Morales, and
  Ortega-Garcia]{tolosana2020deepfakes}
Ruben Tolosana, Ruben Vera-Rodriguez, Julian Fierrez, Aythami Morales, and
  Javier Ortega-Garcia.
\newblock Deepfakes and beyond: A survey of face manipulation and fake
  detection.
\newblock \emph{arXiv preprint arXiv:2001.00179}, 2020.

\bibitem[Turner et~al.(2018)Turner, Hung, Saatci, and
  Yosinski]{Turner2018MetropolisHastingsGA}
Ryan~C Turner, Jane Hung, Yunus Saatci, and Jason Yosinski.
\newblock Metropolis-hastings generative adversarial networks.
\newblock In \emph{ICML}, 2018.

\bibitem[Wang et~al.(2018)Wang, Yu, Wu, Gu, Liu, Dong, Qiao, and
  Change~Loy]{wang2018esrgan}
Xintao Wang, Ke~Yu, Shixiang Wu, Jinjin Gu, Yihao Liu, Chao Dong, Yu~Qiao, and
  Chen Change~Loy.
\newblock Esrgan: Enhanced super-resolution generative adversarial networks.
\newblock In \emph{Proceedings of the European Conference on Computer Vision
  (ECCV)}, 2018.

\bibitem[Wu et~al.(2019)Wu, Donahue, Balduzzi, Simonyan, and
  Lillicrap]{Wu2019LOGANLO}
Yan Wu, Jeff Donahue, David Balduzzi, Karen Simonyan, and Timothy~P. Lillicrap.
\newblock Logan: Latent optimisation for generative adversarial networks.
\newblock \emph{ArXiv}, abs/1912.00953, 2019.

\bibitem[Xie et~al.(2017)Xie, Girshick, Doll{\'a}r, Tu, and
  He]{xie2017aggregated}
Saining Xie, Ross Girshick, Piotr Doll{\'a}r, Zhuowen Tu, and Kaiming He.
\newblock Aggregated residual transformations for deep neural networks.
\newblock In \emph{Proceedings of the IEEE conference on computer vision and
  pattern recognition}, pages 1492--1500, 2017.

\bibitem[Yu et~al.(2018)Yu, Lin, Yang, Shen, Lu, and Huang]{yu2018generative}
Jiahui Yu, Zhe Lin, Jimei Yang, Xiaohui Shen, Xin Lu, and Thomas~S Huang.
\newblock Generative image inpainting with contextual attention.
\newblock In \emph{Proceedings of the IEEE conference on computer vision and
  pattern recognition}, pages 5505--5514, 2018.

\bibitem[Zhang et~al.(2018{\natexlab{a}})Zhang, Goodfellow, Metaxas, and
  Odena]{zhang2018self}
Han Zhang, Ian Goodfellow, Dimitris Metaxas, and Augustus Odena.
\newblock Self-attention generative adversarial networks.
\newblock \emph{arXiv preprint arXiv:1805.08318}, 2018{\natexlab{a}}.

\bibitem[Zhang et~al.(2018{\natexlab{b}})Zhang, Isola, Efros, Shechtman, and
  Wang]{zhang2018unreasonable}
Richard Zhang, Phillip Isola, Alexei~A Efros, Eli Shechtman, and Oliver Wang.
\newblock The unreasonable effectiveness of deep features as a perceptual
  metric.
\newblock In \emph{Proceedings of the IEEE Conference on Computer Vision and
  Pattern Recognition}, pages 586--595, 2018{\natexlab{b}}.

\bibitem[Zhao et~al.(2020{\natexlab{a}})Zhao, Liu, Lin, Zhu, and
  Han]{zhao2020differentiable}
Shengyu Zhao, Zhijian Liu, Ji~Lin, Jun-Yan Zhu, and Song Han.
\newblock Differentiable augmentation for data-efficient gan training.
\newblock \emph{arXiv preprint arXiv:2006.10738}, 2020{\natexlab{a}}.

\bibitem[Zhao et~al.(2020{\natexlab{b}})Zhao, Li, Yu, Gao, and
  Chen]{zhao2020feature}
Yang Zhao, Chunyuan Li, Ping Yu, Jianfeng Gao, and Changyou Chen.
\newblock Feature quantization improves gan training.
\newblock \emph{arXiv preprint arXiv:2004.02088}, 2020{\natexlab{b}}.

\bibitem[Zhou et~al.(2017)Zhou, Lapedriza, Khosla, Oliva, and
  Torralba]{zhou2017places}
Bolei Zhou, Agata Lapedriza, Aditya Khosla, Aude Oliva, and Antonio Torralba.
\newblock Places: A 10 million image database for scene recognition.
\newblock \emph{IEEE Transactions on Pattern Analysis and Machine
  Intelligence}, 2017.

\bibitem[Zhu et~al.(2020)Zhu, Shen, Zhao, and Zhou]{zhu2020domain}
Jiapeng Zhu, Yujun Shen, Deli Zhao, and Bolei Zhou.
\newblock In-domain gan inversion for real image editing.
\newblock \emph{arXiv preprint arXiv:2004.00049}, 2020.

\end{thebibliography}
}

\newpage
\appendix
\section{Detailed Description of Evaluation Metrics}
\label{sec:evlaution_metric_description}
We use a variety of evaluation metrics to diagnose the effect that training with instance selection has on the learned distribution. In all cases where a reference distribution is required we use \emph{the original training distribution}, and not the distribution produced after instance selection. Doing so would unfairly favour the evaluation of instance selection, since the reference distribution could be changed to one that is trivially easy to generate.

\textbf{Inception Score (IS)}~\citep{salimans2016improved} evaluates samples by extracting class probabilities from an ImageNet pretrained Inceptionv3 classifier and measuring the distribution of outputs over all samples. The Inception Score is maximized when a model produces highly recognizable outputs for each of the ImageNet classes. One of the major limitations of the Inception Score is its insensitivity to mode collapse within each class. A model that produces a single high quality image for each category can still achieve a good score.
 
\textbf{Fr\'echet Inception Distance (FID)}~\citep{heusel2017gans} measures the distance between a generated distribution and a reference distribution, as approximated by a Gaussian fit to samples projected into the feature space of a pretrained Inceptionv3 model. FID has been shown to correlate well with image quality, and is capable of detecting mode collapse and mode adding. However,  FID does not differentiate between fidelity and diversity. As such, it is difficult to assess whether a model has achieved a good FID score based on good mode coverage, or because it produces high quality samples.

\textbf{Precision and Recall (P\&R)}~\citep{kynkaanniemi2019improved} were designed to address the limitations of FID by providing separate metrics to evaluate fidelity and diversity. To calculate P\&R, image manifolds are created by first embedding each image in a given distribution into the feature space of a pretrained classifier. A radius is then extended from each data point to its $K$\textsuperscript{th} nearest neighbour to form a hypersphere, and the union of all hyperspheres represents the image manifold. Precision is  described as the percentage of generated samples that fall within the manifold of real images. Recall is described as the percentage of real images which fall within the manifold of generated samples. A limitation of P\&R is that they are susceptible to outliers, both in the reference and generated distributions~\citep{naeem2020reliable}. Outliers artificially inflate the size of the image manifolds, increasing the rate at which samples fall into those manifolds. Thus, a dataset or model that produces many outliers may achieve scores that are better than the quality of the samples would indicate.

\textbf{Density and Coverage (D\&C)}~\citep{naeem2020reliable} have recently been proposed as robust alternatives to Precision and Recall. Density can be seen as an extension of Precision which measures how many real image manifolds a generated sample falls within on average. Coverage is described as the percentage of real images that have a generated sample fall within their manifold. 

\textbf{Classification Accuracy Score (CAS)}~\citep{ravuri2019classification,shmelkov2018good} was introduced for evaluating the usefulness of conditional generative models for augmenting downstream tasks such as image classification. To compute CAS, generated samples are used to train a classifier, which is then used to classify real data from a test set. Generally, it is observed that models with greater sample diversity achieve higher CAS, with image fidelity being of less importance. We do not evaluate CAS for the majority of our experiments as it is very computationally expensive to compute, but we do report it in \S~\ref{sec:cas_128}, Table \ref{table:cas} for our 128 \texttimes\ 128 ImageNet BigGAN experiments as a reference for how instance selection affects CAS.

\section{Additional Evaluation Metrics - Classification Accuracy Score (CAS)}
\label{sec:cas_128}

We compute CAS by training a ResNet50 on samples from each of our 128 \texttimes\ 128 BigGAN models using the standard ImageNet pipeline from PyTorch\footnote{\url{https://github.com/pytorch/examples/tree/master/imagenet}}. We find that the model trained without instance selection achieves the best CAS, which is expected given that this model also produces more diverse samples (as measured by Recall). Interestingly, CAS for the BigGAN trained with instance selection drops by less than 1\%, despite it only having seen 50\% of the ImageNet training set. This result might suggest that neither of the models evaluated does a good job at generating recognizable outliers from the ImageNet training set.

\begin{table}[h]
\centering
\caption{CAS for BigGAN trained with and without instance selection. Following \citep{ravuri2019classification}, both models use a truncation ratio of 1.5 when generating samples for increased diversity.}
\label{table:cas}
\begin{tabular}{lccc}
\toprule
Training Set & Resolution & Top-5 Accuracy & Top-1 Accuracy \\
\midrule
BigGAN & $128\times128$ & \textbf{18.73} & \textbf{9.21} \\
BigGAN + 50\% inst. sel. & $128\times128$ & 17.94 & 8.42 \\
\bottomrule
\end{tabular}
\end{table}

\section{Scores of Evaluation Metrics on Real Data}
\label{sec:scores_on_real_data}
For each evaluation metric we compute scores on real data (Table \ref{table:real_data}) as a reference for comparison with the values produced by generative models. These values can be thought of as the scores which would be achieved by a generative model that perfectly captures the target distribution. Metrics are evaluated on the ImageNet validation set, using all 50k data points for IS and FID and 10k randomly selected data points for P\&R and D\&C. Note that it is possible for generative models to surpass the scores of real data for metrics that focus on image fidelity, such as IS, P, and D, but these models often have proportionally lower diversity scores.

\begin{table}[h]
\centering
\caption{Scores of real data from the ImageNet validation set for all evaluation metrics.}
\label{table:real_data}
\begin{tabular}{l|c|c|cc|cc}
\toprule
Resolution & IS $\uparrow$  & FID $\downarrow$ & P $\uparrow$  & R $\uparrow$  & D $\uparrow$  & C $\uparrow$ \\
\midrule
64 \texttimes\ 64 & 59.1 & 1.0 & 0.79 & 0.79 & 0.99 & 0.96 \\
128 \texttimes\ 128 & 148.2 & 1.2 & 0.84 & 0.82 & 1.01 & 0.96 \\
256 \texttimes\ 256 & 225.9 & 1.4 & 0.85 & 0.83 & 1.01 & 0.96 \\
\bottomrule
\end{tabular}
\end{table}

\section{Retention Ratio Experiment Numerical Results}

In Table \ref{table:retention_ratio_sweep_results} we include numerical results for the retention ratio experiments conducted in \S \ref{sec:retention_rate}. These values accompany the plots in Figure \ref{fig:keep_percentage_plot}. We also report the performance of BigGAN and FQ-BigGAN from \cite{zhao2020feature} for comparison.

\begin{table}[h]
\centering
\caption{Performance of models trained on $64\times64$ resolution ImageNet. A retention ratio of less than 100 indicates that instance selection is used. Best results in bold.}
\label{table:retention_ratio_sweep_results}
\begin{tabular}{lccc|c|c|cc|cc}
\toprule
Model &\begin{tabular}[c]{@{}c@{} }Params \\ (M)\end{tabular} & \begin{tabular}[c]{@{}c@{} }Batch \\ Size\end{tabular} & \begin{tabular}[c]{@{}c@{} }Retention \\ Ratio (\%)\end{tabular}& IS $\uparrow$ & FID $\downarrow$ & P $\uparrow$ & R $\uparrow$ & D $\uparrow$ & C $\uparrow$\\
\midrule
BigGAN & 52.54 & 512 & 100 & 25.43 & 10.55 & - & - & - & - \\
FQ-BigGAN & 52.54 & 512 & 100 & 25.96 & 9.67 & - & - & - & - \\
\midrule
\multirow{10}{*}{SAGAN} & \multirow{10}{*}{23.64} & \multirow{10}{*}{128} & 100 & 17.77 & 17.23 & 0.68 & \textbf{0.66} & 0.72 & 0.71 \\
 & & & 90 & 18.98 & 15.85 & 0.70 & \textbf{0.66} & 0.75 & 0.74 \\
 & & & 80 & 21.62 & 13.17 & 0.74 & 0.65 & 0.87 & 0.79 \\
 & & & 70 & 23.95 & 11.98 & 0.75 & 0.64 & 0.92 & 0.82 \\
 & & & 60 & 27.95 & 10.35 & 0.78 & 0.63 & 0.99 & 0.87 \\
 & & & 50 & 31.04 & 9.63 & 0.79 & 0.62 & 1.07 & 0.88 \\
 & & & 40 & 37.10 & \textbf{9.07} & 0.81 & 0.60 & 1.12 & \textbf{0.90} \\
 & & & 30 & 41.85 & 9.75 & \textbf{0.83} & 0.55 & \textbf{1.19} & \textbf{0.90} \\
 & & & 20 & \textbf{43.30} & 12.36 & 0.82 & 0.49 & 1.17 & 0.88 \\
 & & & 10 & 37.16 & 19.24 & 0.79 & 0.33 & 1.07 & 0.78 \\
\bottomrule
\end{tabular}
\end{table}

\section{Complementarity of Instance Selection and Truncation}
\label{sec:truncation_analysis}

The truncation trick is a simple and popular technique which is used to increase the visual fidelity of samples from a GAN at the expense of reduced diversity~\citep{brock2018large}. This trade-off is achieved by biasing latent samples towards the interior regions of the latent distribution, either by truncating the distribution, or by interpolating latent samples towards the mean~\citep{karras2019style,kynkaanniemi2019improved}. 

\begin{figure}[h]
    \centering
    \includegraphics[width=0.93\textwidth]{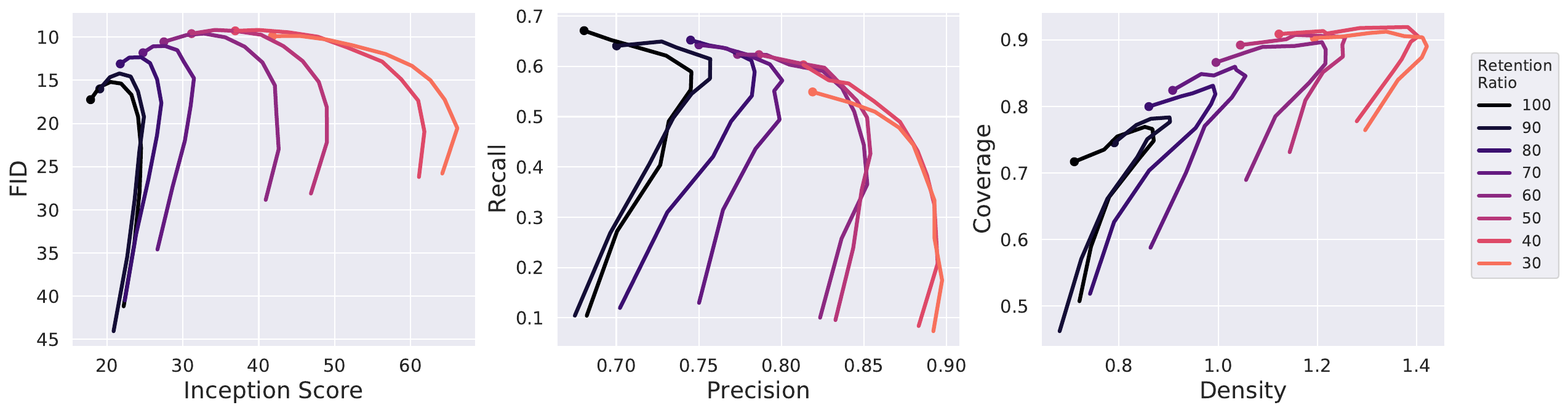}
    \caption{Truncation trick applied to models trained with instance selection for truncation thresholds 1 to 0.1. The base models (threshold = 1) are marked with a $\bullet$. Up and to the right is best.}
    \label{fig:keep_percentage_truncation_sweep}
\end{figure}

To examine the compatability between the truncation trick and instance selection, we truncate latent vectors of the models trained in \S \ref{sec:retention_rate}, varying the truncation threshold from 1.0 to 0.1 (Figure \ref{fig:keep_percentage_truncation_sweep}). We observe that combining both techniques results in a greater improvement in visual fidelity than either method applied in isolation. We anticipate that other post-hoc filtering methods could also see complimentary benefits when combined with instance selection, such as DRS, MH-GAN, and DDLS.

\section{Insights for Applying Instance Selection to GANs}
We found that, while instance selection could be used to achieve significant gains in model performance, some changes to other hyperparameters were necessary in order to ensure training stability. Here we detail some techniques that we found to work well in our experiments.

\begin{itemize}
    \item \textbf{Reduce batch size} - Contrary to evidence from BigGAN~\citep{brock2018large} suggesting that larger batch sizes improve GAN performance, we found batch sizes larger than 256 to degrade performance when training with instance selection. We speculate that because we have simplified the training distribution by removing the difficult examples, the discriminator overfits the training set much faster. We posit that the smaller batch size could be acting as a form of regularization by reducing the accuracy of the gradients, thereby allowing the generator to train for longer before the discriminator overfits the training set and the model collapses. 
    \item \textbf{Reduce model capacity} - Since the complexity of the training set is reduced when applying instance selection, we found it necessary in some cases to also reduce model capacity. Training models with too much capacity lead to early collapse, also likely caused by the discriminator quickly overfitting the training set. We note that with proper regularization, models trained with instance selection could still benefit from more capacity.
    \item \textbf{Apply additional regularization} - We have not experimented much with applying GAN regularization methods to our models, but think that it could be important for combating the aforementioned discriminator overfitting problem. Applying techniques such as R1 regularization~\citep{mescheder2018training} or recently proposed GAN data augmentation~\citep{karras2020training, zhao2020differentiable} could allow for instance selection to be combined with the benefits of larger batch sizes and model capacity. We leave this investigation for future work.
\end{itemize}

\section{Sample Sheets}
\label{sec:sample_sheets}

We generate several different sample visualizations in order to better understand the impact that instance selection has on GAN behaviour. 

\begin{figure}[h]
    \centering
    \includegraphics[width=0.85\textwidth]{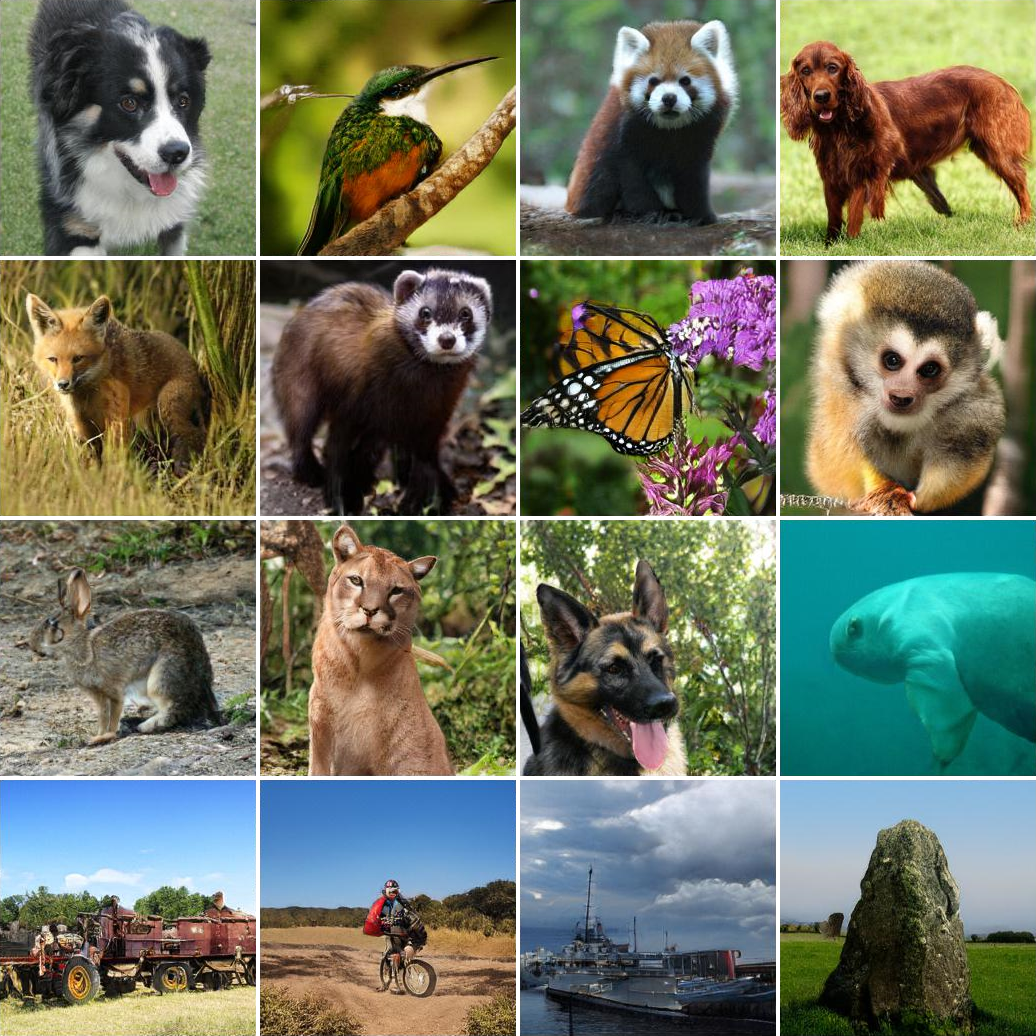}
    \caption{Photorealistic samples from BigGAN trained on $256 \times 256$ ImageNet with 50\% instance selection. Samples are manually selected to showcase the best quality outputs from this model.}
    \label{fig:photorealistic_256x256}
\end{figure}

In Figure \ref{fig:photorealistic_256x256} we showcase some photorealistic samples generated by a $256 \times 256$ BigGAN model trained with instance selection.

\begin{figure}[h]
     \centering
     \begin{subfigure}[b]{0.44\textwidth}
         \centering
         \includegraphics[width=\textwidth]{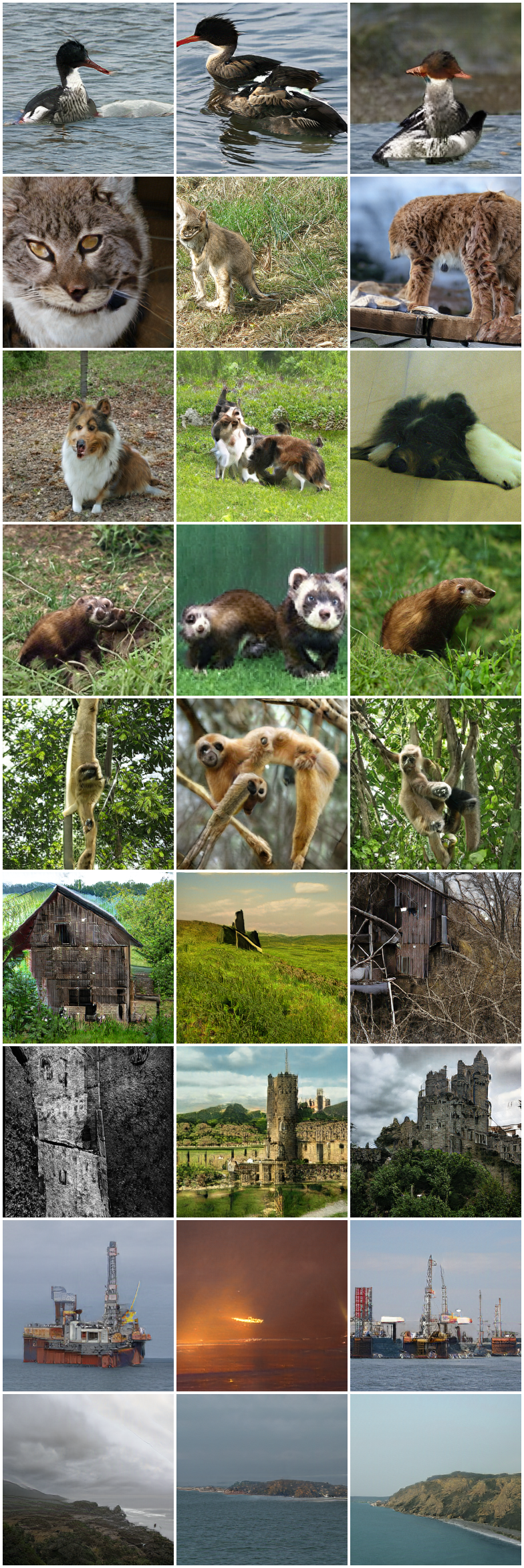}
         \caption{Baseline ($100\%$ of dataset)}
         \label{fig:imagenet_random_classes_baseline}
     \end{subfigure}
     \quad
     \begin{subfigure}[b]{0.44\textwidth}
         \centering
         \includegraphics[width=\textwidth]{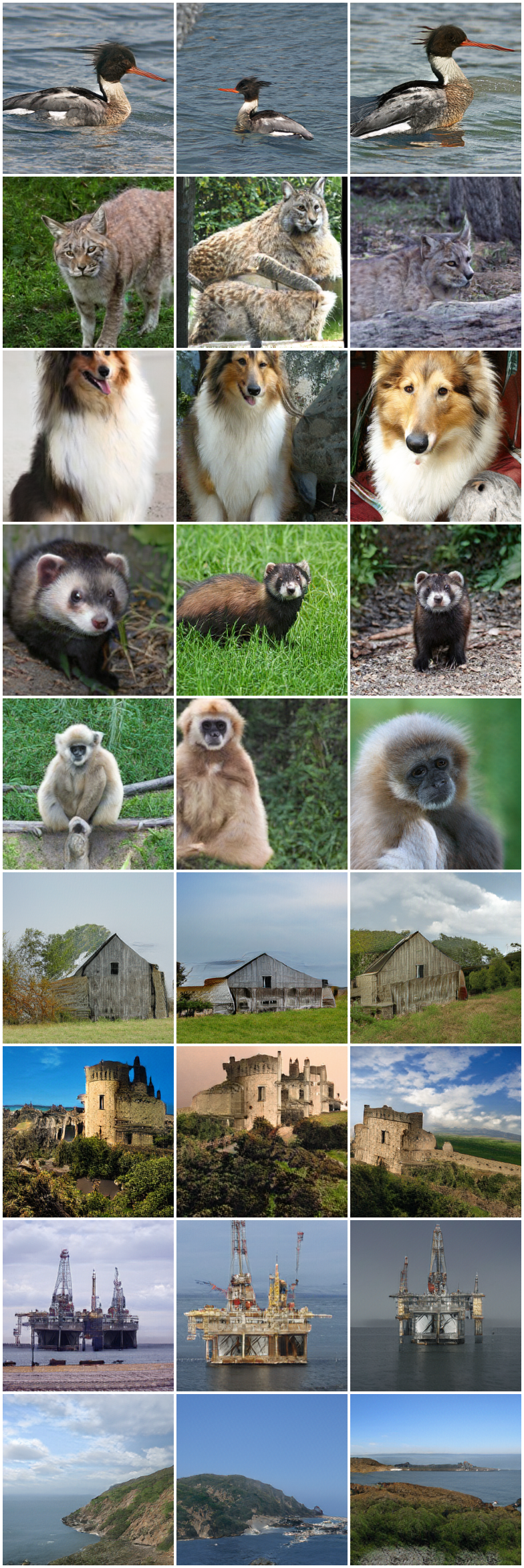}
         \caption{w\textbackslash ~ Instance selection ($50\%$ of dataset)}
         \label{fig:imagenet_random_classes_reduced_gan}
     \end{subfigure}
        \caption{Uncurated samples from BigGAN models trained on 256 \texttimes 256 resolution ImageNet. Each row is conditioned on a different class (from top): Red-breasted Merganser, Lynx, Collie, Mink, Gibbon, Barn, Castle, Drilling Platform, Promontory.}
        \label{fig:imagenet_256_random_class_samples}
\end{figure}

In Figure \ref{fig:imagenet_256_random_class_samples} we compare randomly selected samples from the official pretrained $256 \times 256$ BigGAN (Figure \ref{fig:imagenet_random_classes_baseline}) with random samples from our $256 \times 256$ BigGAN trained with 50\% instance selection (Figure \ref{fig:imagenet_random_classes_reduced_gan}). Samples from the instance selection model appear more realistic on average.

\begin{figure}[h]
     \centering
     \begin{subfigure}[b]{0.48\textwidth}
         \centering
         \includegraphics[width=\textwidth]{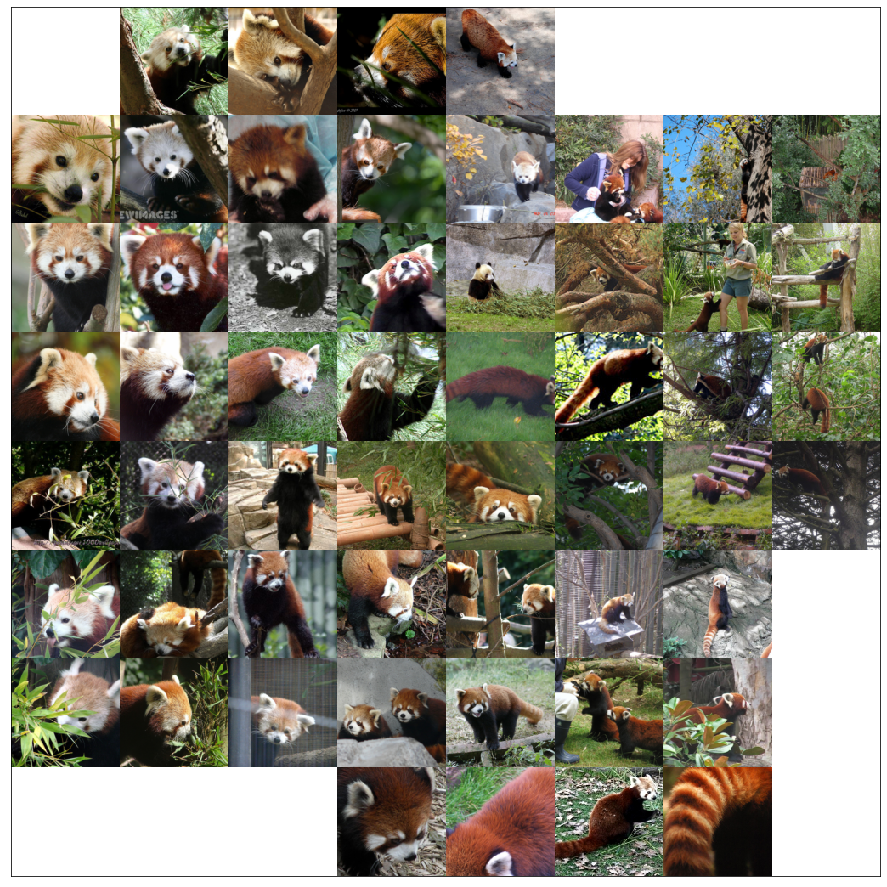}
         \caption{Full dataset}
         \label{fig:umap_original_dataset}
     \end{subfigure}
     \quad
     \begin{subfigure}[b]{0.48\textwidth}
         \centering
         \includegraphics[width=\textwidth]{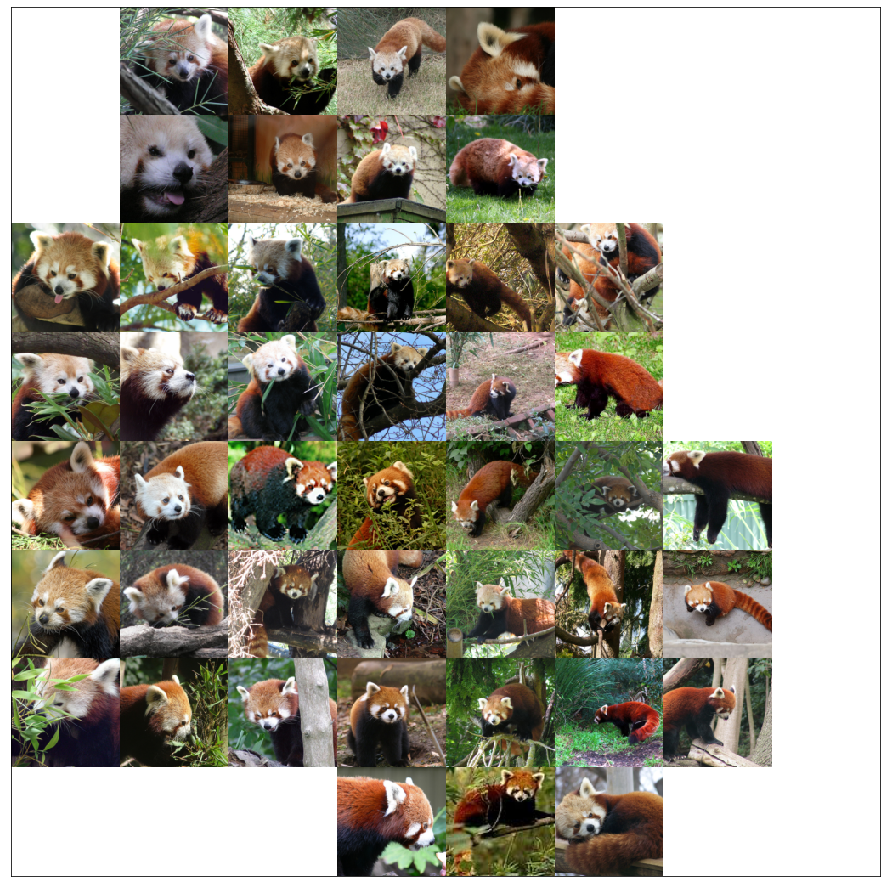}
         \caption{Dataset after $50\%$ instance selection}
         \label{fig:umap_reduced_dataset}
     \end{subfigure}
     \begin{subfigure}[b]{0.48\textwidth}
         \centering
         \includegraphics[width=\textwidth]{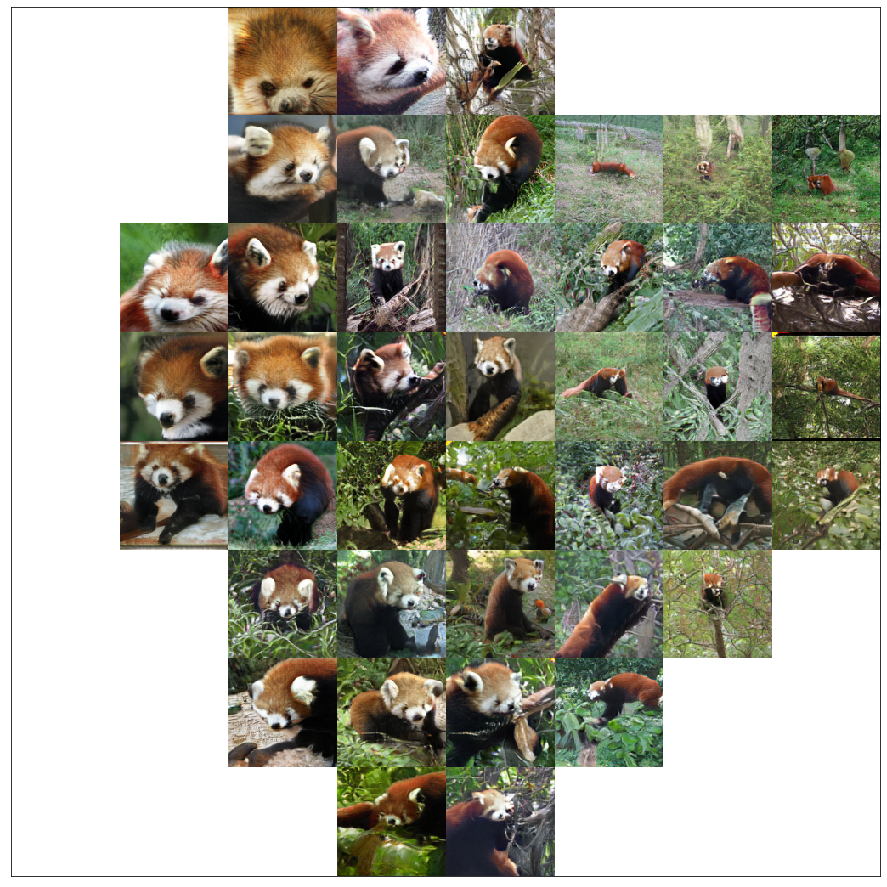}
         \caption{Samples from GAN trained on full dataset}
         \label{fig:umap_baseline_gan}
     \end{subfigure}
     \quad
     \begin{subfigure}[b]{0.48\textwidth}
         \centering
         \includegraphics[width=\textwidth]{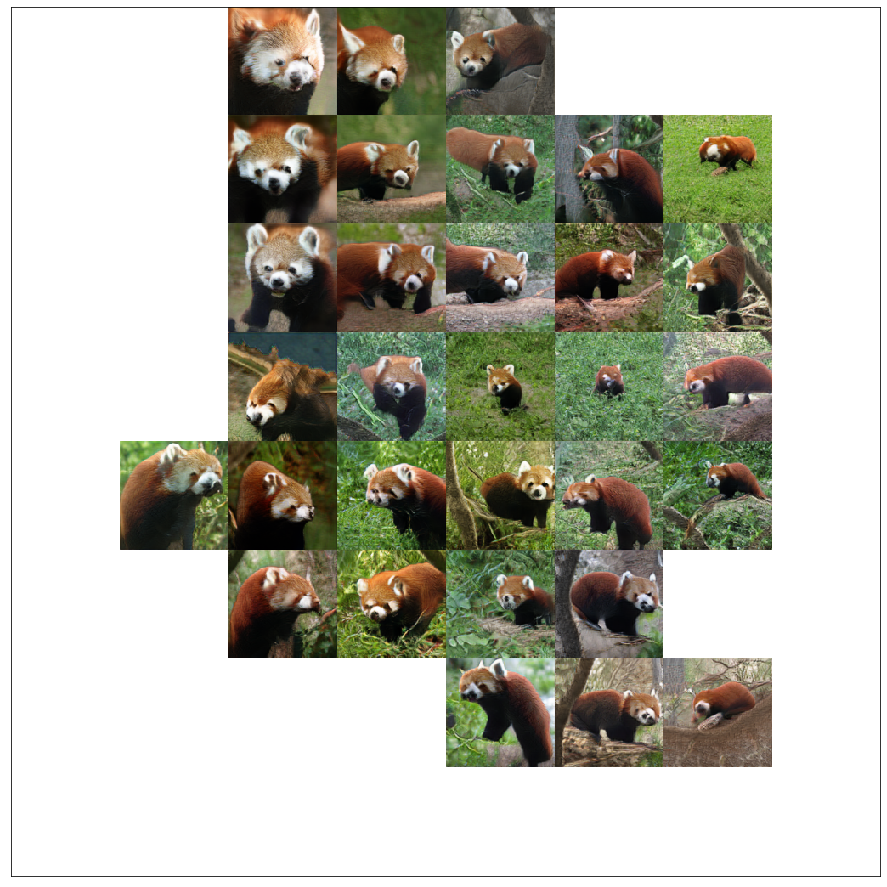}
         \caption{Samples from GAN trained on $50\%$ of dataset}
         \label{fig:umap_reduced_gan}
     \end{subfigure}

    \caption{Visualization of the image manifolds for the red pandas class from (a) the full ImageNet dataset, (b) the dataset after $50\%$ instance selection, (c) samples from a GAN trained on the full dataset, and (d) samples from a GAN trained on $50\%$ of the dataset. All images are at $128 \times 128$ resolution. Manifolds are created by embedding all images into an Inceptionv3 feature space, then projecting them into 2D with UMAP~\citep{mcinnes2018umap}. All images share the same 2D embedding such that subplots are comparable. Instance selection removes images from the dataset that have unusual viewpoints or pose. Both GANs appear to cover less of the image manifold than their respective source datasets. The GAN trained on the full dataset covers some regions of the image manifold that are not covered by the model trained with instance selection, however, these regions are more likely to appear unrealistic.}
    \label{fig:umaps}
\end{figure}

To better understand how instance selection affects sample diversity, we visualize image manifolds of different datasets and models by organizing images in 2D using UMAP~\citep{mcinnes2018umap} (Figure \ref{fig:umaps}). We only plot a single class so that we can see variations across the image manifold in greater detail than if multiple classes were plotted simultaneously.
All image samples share the same 2D embedding, such that manifolds are comparable between datasets and models. We observe that even though instance selection has removed $50\%$ of the images from the original dataset (Figure \ref{fig:umap_original_dataset}), it still retains coverage over most of the original image manifold (Figure \ref{fig:umap_reduced_dataset}). Only images containing extreme viewpoints are omitted. The GANs trained on the original and reduced datasets both cover less of the image manifold than their respective source datasets. While the baseline GAN (Figure \ref{fig:umap_baseline_gan}) covers more of the image manifold than the GAN trained with instance selection (Figure \ref{fig:umap_reduced_gan}), samples from these extra regions often appear less realistic.

\end{document}